\documentclass{article}
\usepackage{spconf,amsmath,graphicx}
\usepackage{amsfonts,amssymb}
\usepackage{xcolor,colortbl}
\usepackage{paralist}
\usepackage{booktabs}

\definecolor{Gray}{gray}{0.85}
\definecolor{LightCyan}{rgb}{0.88,1,1}

\newcolumntype{a}{>{\columncolor{Gray}}c}
\newcolumntype{b}{>{\columncolor{white}}c}

\title{Flash Finds Gray Pixels}
\title{Flash Photography Gray Pixel}
\title{Flash Gray Pixel}
\title{Flash Lightens Gray Pixels}
\name{Yanlin Qian$^{\star}$, Song Yan$^{ \star}$, Joni-Kristian K\"am\"ar\"ainen$^{\star}$, Jiri Matas$^{ \dagger}$}
\address{
  $^\star$Computing Sciences, Tampere University\\
  $^\dagger$Center for Machine Perception, Czech Technical University in Prague\\
}
\begin{document}
%
\maketitle
\begin{abstract}

In the real world, a scene is usually cast by multiple illuminants and herein we address the problem of spatial illumination estimation. Our solution is based on detecting gray pixels with the help of flash photography. We show that flash photography significantly improves the performance of gray pixel detection without
illuminant prior, training data or calibration of the flash. We also introduce a novel flash photography
dataset generated from the MIT intrinsic dataset.
\end{abstract}

\section{Introduction}
\label{sec:intro}

We address the \textit{illumination estimation} problem which aims to measure the chroma of illumination in order to remove the color-bias from a captured image~\cite{finlayson2018colour}.  Illumination estimation can help in high-level vision tasks, \textit{e.g.} object recognition, tracking \cite{foster2011color} and intrinsic image decomposition. There exists a large number of related works, from the
traditional non-learning approaches~\cite{brainard1986analysis} to recent deep learning based
approaches~\cite{yanlin2016icpr,hu2017cvpr,yanlin2017iccv}. However, the vast majority of these
works concentrate on the case of a single global illumination which is often an invalid assumption~\cite{hui2016white}. In this paper, we explore a more-complex less-optimistic setting -- mixed illumination~\footnote{We traverse related works and there exists multiple terms referring to the same thing (which may confuse readers), i.e.~mixed illumination, spatially-varying illumination, multiple illumination, mixed lighting condition.}. 

Spatially-varying illumination refers to that on a captured scene, each pixel captures different number of light phantoms when the camera shutter is on. In other words, all pixels do not share the same configuration of lights~ \cite{hui2016white}, which is the default assumption for single-illumination estimation. Compared to the single global illumination setting, the mixed illumination setting better corresponds to the real world \cite{joze2014exemplar}, but is more challenging clearly, as it extends the ill-posed problem from one point to a spatial map  \cite{hui2016white}, without extra knowledge or input.

To circumvent the hardness that spatially-varying illumination brings, efforts are put as follows: user guidance or human interaction is given as a supervisory signal \cite{lischinski2006interactive,boyadzhiev2012user}; small patch is assumed to be cast by only one light \cite{riess2011illuminant}; light color and number need to be known before experiments \cite{hsu2008light}. Unlike these methods, we make use of flash photography.

\setlength{\tabcolsep}{1pt}
\renewcommand{\arraystretch}{1}
\begin{figure}
\begin{center}

\begin{tabular}{c ccc cc}

\hspace{-15pt}
\vspace{-0.75mm}
\raisebox{2\height}{\rotatebox{0}{{ ambi }}}
&{{\includegraphics[width=1.6cm,height=1cm]{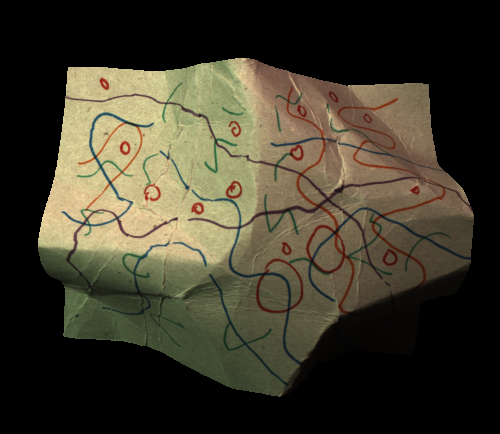}}}
& 
&{{\includegraphics[width=1.6cm,height=1cm]{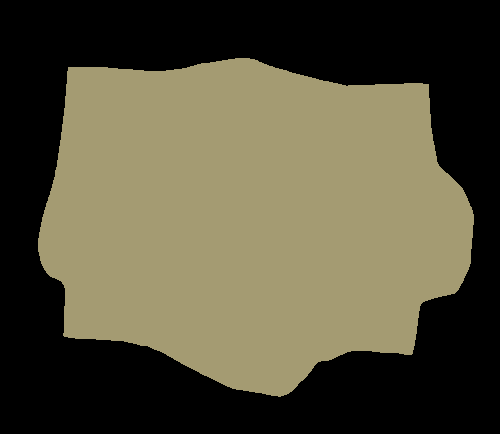}}}
&
&{{\includegraphics[width=1.6cm,height=1cm]{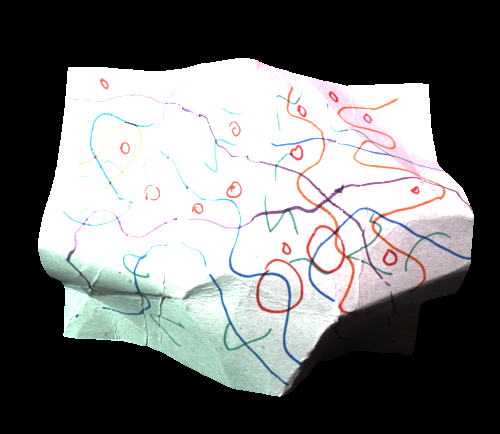}}}
\\

\hspace{-15pt}
\vspace{-0.75mm}
\raisebox{2\height}{\rotatebox{0}{{ +flash }}}
&{{\includegraphics[width=1.6cm,height=1cm]{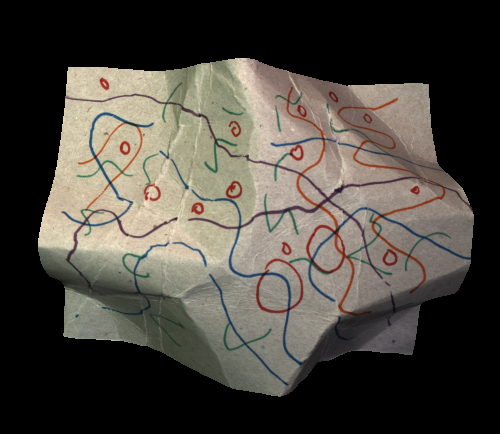}}}
&{{\includegraphics[width=1.6cm,height=1cm]{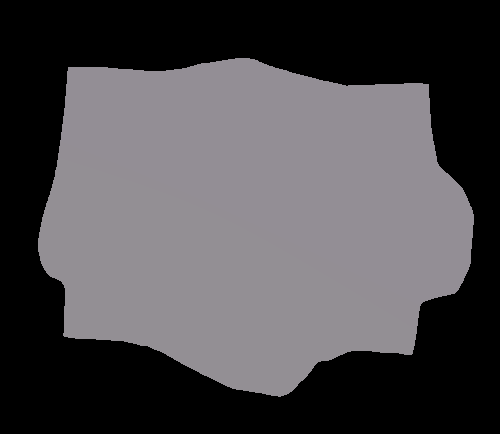}}}
&{{\includegraphics[width=1.6cm,height=1cm]{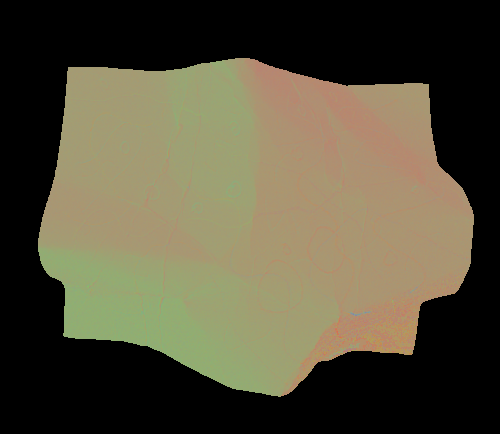}}}
&{{\includegraphics[width=1.6cm,height=1cm]{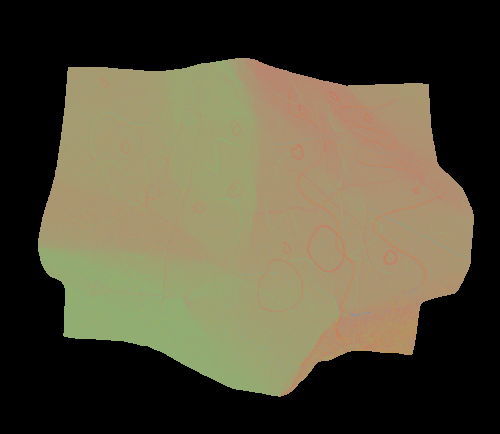}}}
&{{\includegraphics[width=1.6cm,height=1cm]{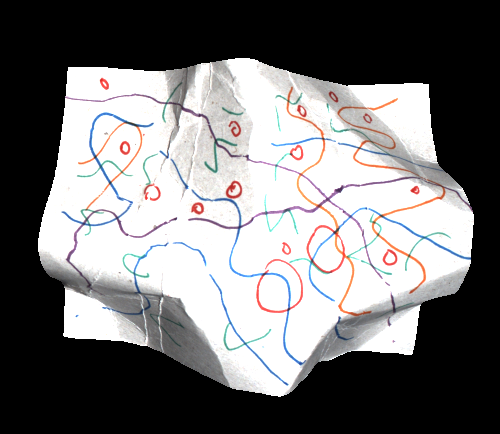}}}
\\

\vspace{-3mm}
\raisebox{2\height}{\rotatebox{0}{{  }}}
& \raisebox{2\height}{\textit{img}}
& \raisebox{2\height}{\textit{pred\_flash}}
& \raisebox{2\height}{\textit{pred}}
& \raisebox{2\height}{\textit{GT}}
& \raisebox{2\height}{\textit{corr\_img}}

\end{tabular}

\end{center}
\vspace{-2.5mm}
\caption{
A multi-illuminant image (\textit{ambi}), Gray Pixel~\cite{yang2015efficient} outputs an erroneous prediction (single value, \textit{pred}) compared to the groundtruth (\textit{GT}). By virtue of the flash image (\textit{+flash}) the proposed
flash photography gray pixel provides an accurate spatial estimate (\textit{pred}) and faithful corrected image (\textit{corr\_img}). 
}
\label{figure:intro}
\vspace{-3mm}
\end{figure}

Flash photography refers to image processing techniques which use
non-flash/flash image pairs.
This technique is well adopted to spatial illumination estimation \cite{dicarlo2001illuminating, petschnigg2004digital}, which assume each patch illuminated by one light and obtain decent results. 
Hui~\textit{et al.}~\cite{hui2016white} proposed a closed-form solution of spatial illumination for the case of a calibrated flash.
In essence, flash calibration in~\cite{hui2016white} equals to knowing the ``groundtruth'' surface albedo in a flash-only image. What's more, flash may appear in other forms, \textit{i.e.} varying sunlight, cast shadow, which may be hard to calibrate.

In this paper, we propose a novel spatial illumination estimation method, using flash photography, \textbf{without} need of flash calibration and any other illuminant prior. 
Our method relies on gray pixel detection~\cite{yang2015efficient}. The original work assumes Lambertian surfaces, and then revisited and improved by \cite{yanlin2019vissap,yanlin2019arxiv}. 
The original and extended gray pixel methods however
fail in the case of mixed illumination (the top row in Fig.~\ref{figure:intro}), but we analytically show how flash photography circumvents the problem and ``lightens'' gray pixel photometrically and in performance (bottom row in Fig.~\ref{figure:intro}). 
 The interplay of flash photography and gray pixel enables us to achieve a largely increased performance on our synthesized laboratory dataset and some real-world images, than running gray pixel methods alone, without knowing the flash color.

\vspace{\medskipamount}\noindent\textbf{Our contributions} are three-fold:
\begin{compactitem}
\item We revisit and revise the gray pixel methods for the case of mixed illumination.
\item Leveraging flash photography and the gray pixel method, we propose a novel learning-free and well-performing method for spatial illumination estimation.
\item We propose a novel flash-photography dataset for benchmarking multi-illuminant methods. The dataset is based on the MIT intrinsic dataset. 
\end{compactitem}

The rest of this paper is organized as follows. Section $2$ revisits gray pixel and its variants before we introduce the flash-photography gray pixel in Section $3$. In Section $4$ we describe the MIT intrinsic based dataset for our task. Section $5$ covers the experiments and results. We conclude in Section~$6$.

\section{Gray Pixel}
\label{sec:gp}

Assuming one light source and narrow sensor response, Gray Pixel \cite{yang2015efficient} is derived from the Lambertian model, given as:
\begin{align}
\label{eq:onelight_formation}
I^{c}(p) = R^c(p) \max(n(p)^\intercal s,0)~l^c,
\end{align}
which shows the color channel $c$ at the location $p$ in image $I$ is a function of a surface albedo $R$, surface normal $n$, light direction $s$ and illumination color $l$. 

Following the procedure in \cite{yang2015efficient}, applying \textit{log} and a Mexican hat filter $\delta$ on Eq.~\ref{eq:onelight_formation} yields: 
\begin{align}
\label{eq:gp_formation}
\delta\log I^{c}(p)&=\delta\log R^c(p)+\delta\log \max(n(p)^\intercal s,0)+\delta\log l^c.
\end{align}
A single light casting a small local neighborhood (the same color and direction), Eq.~\ref{eq:gp_formation} simplifies to:
\begin{align}
\label{eq:gp_formation2}
\delta\log I^{c}(p)&=\delta\log R^c(p),
\end{align}
which is the core of gray pixel. $\delta\log I^{c}(p)=\delta\log I^{c'}(p)$, $\forall c,c' \in \{R,G,B\}$ defines a "pure gray pixel". To rank pixels \textit{w.r.t.} ``grayness'', \cite{yang2015efficient} defines the following grayness function, up to a scale:
\begin{align}
\label{eq:gp_formation3}
g(p)=\sum_{c \in R,G,B} (\delta\log I^{c}(p) - \bar\delta\log I(p) )^2 / \bar\delta\log I(p),
\end{align}
where $\bar\delta\log I(p)$ is the mean value of $\delta\log I^{c}(p)$. This method works robustly with single-illumination scenes where diffuse reflection (the
Lambertian assumption) is dominant.

Then \cite{yanlin2019vissap} augments the above Gray Pixel by replacing Eq.~\ref{eq:gp_formation3} with the following luminance-independent function:
\begin{align}
\label{eq:msgp_formation1}
g'(p) =\cos^{-1} \left(\frac{1}{\sqrt{3}} \frac{\Vert{ \delta \log I(p)}\Vert_1}
{\Vert  \delta \log I(p) \Vert_2}  \right),
\end{align}
where $\Vert \cdot \Vert_n$ refers to the $\ell n$ norm. $g'(p)=0$ refers to pure gray pixel. To remove spurious color pixels, \cite{yanlin2019vissap} applies a mean shift clustering to choose the strongest mode -- dominant illumination vector.

The mechanism to detect gray pixel is further improved in \cite{yanlin2019arxiv}. Different to \cite{yang2015efficient,yanlin2019vissap}, which are based on the Lambertian model, Qian~\textit{et al.}~\cite{yanlin2019arxiv} uses the {\em dichromatic reflection model}~\cite{shafer1985using} to derive a set of more strict constraints for gray pixels: 
\begin{align}
\label{eq:dgp_formation1}
g''(p) = \Vert \delta(\log(I^R)-\log(\vert I \vert)),
\delta(\log(I^G)-\log(\vert I \vert))\Vert_2,
\end{align}
where $\vert I \vert$ refers to $(I^R+I^G+I^B)$. We refer readers to the original paper for more details. In Section \ref{sec:results} we report their performance in a mixed-lighting dataset and show how flash photography improves them.


\section{Flash Gray Pixel}
\label{sec:flash_gp}

In the sequel, we first investigate what will happen to the existing gray pixel methods in the
case of mixed illumination. Then we propose a novel gray pixel method using flash photography, termed as \textit{Flash Gray Pixel}.

Here we do analysis on the original Gray Pixel \cite{yang2015efficient}, but similar conclusion for \cite{yanlin2019vissap,yanlin2019arxiv} can be inferred in an analogue
manner. A scene is illuminated by $N$ light sources or arbitrary type and color.
To describe the image formation process in this case,
Eq.~\ref{eq:onelight_formation} is modified to:
\begin{align}
\label{eq:morelight_formation}
I^{c}(p) = R^c(p) \sum_i \lambda_i(p)~l_i^c \enspace ,
\end{align}
where $\lambda_i(p)$ represents the shading term $\max(n(p)^\intercal s_i,0)$. Eq.~\ref{eq:gp_formation2} changes to: 
\begin{align}
\label{eq:gp_formation2_morelight}
\delta\log I^{c}(p)&=\delta\log R^c(p) + \delta\log(\sum_i \lambda_i(p)~l_i^c) \enspace .
\end{align}
Since that is a mixed illumination image, the light configuration varies from pixel to pixel, making the right most term in Eq.~\ref{eq:gp_formation2_morelight} non-zero and therefore Eq.~\ref{eq:gp_formation2} fails. This finding explains that mixed illumination hinders the performance of the gray pixel method, which motivates us to leverage flash photography. 

When a flash light is present, the flash image $I_f$ is expressed as: 
\begin{align}
\label{eq:flashimage_formation}
I_f^{c}(p) = R^c(p) (\sum_i \lambda_i(p)~l_i^c + \lambda_f(p)~l_f^c),
\end{align}
where $\lambda_f(p)$ is the shading term of flash light at the position $p$ and $l_f$ the unit-norm chroma vector. Subtracting Eq.~\ref{eq:morelight_formation} from Eq.~\ref{eq:flashimage_formation}, we get a {\em flash-only image} $I_\textit{fo}$:
\begin{align}
\label{eq:flashimage_formation2}
I_\textit{fo}^{c}(p) = R^c(p) \lambda_f(p)~l_f^c,
\end{align}
which is a more solid ground for searching gray pixels. In other terms,
flash image helps to remove the negative effect of spatially-varying light configurations and gray pixels are now {\em flash gray pixels}.

\setlength{\tabcolsep}{1pt}
\renewcommand{\arraystretch}{1}
\begin{figure*}
\begin{center}

\begin{tabular}{c ccc ccc}

\vspace{-0.75mm}
\raisebox{2\height}{\rotatebox{0}{{ 2 }}}
&{{\includegraphics[width=1.8cm,height=1.2cm]{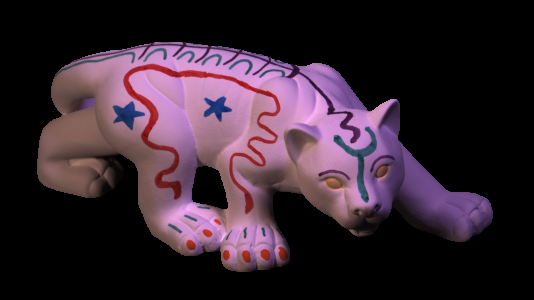}}}
&{{\includegraphics[width=1.8cm,height=1.2cm]{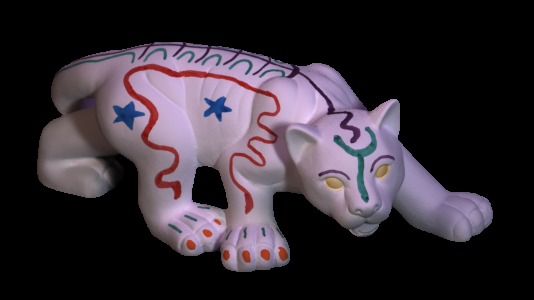}}}
&{{\includegraphics[width=1.8cm,height=1.2cm]{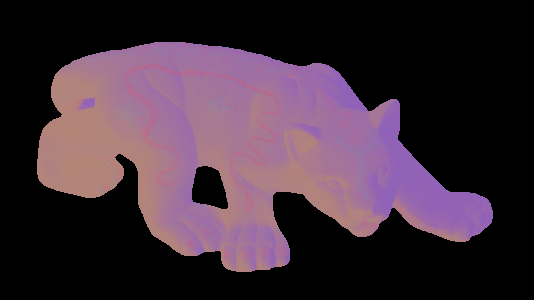}}}
&{{\includegraphics[width=1.8cm,height=1.2cm]{2illu/paper2_ambient.png}}}
&{{\includegraphics[width=1.8cm,height=1.2cm]{2illu/paper2_flash.png}}}
&{{\includegraphics[width=1.8cm,height=1.2cm]{2illu/paper2_illu.png}}}
\\
\vspace{-0.75mm}
\raisebox{2\height}{\rotatebox{0}{{ 3 }}}
&{{\includegraphics[width=1.8cm,height=1.2cm]{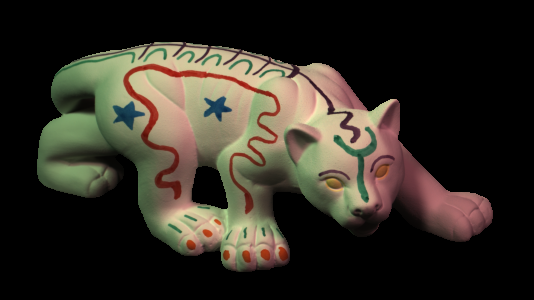}}}
&{{\includegraphics[width=1.8cm,height=1.2cm]{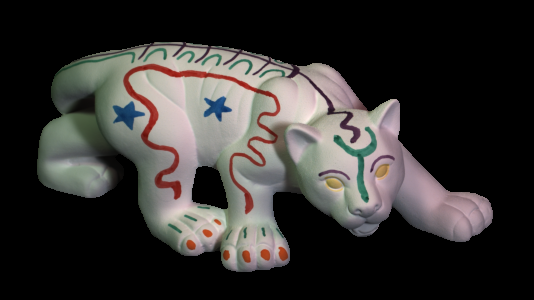}}}
&{{\includegraphics[width=1.8cm,height=1.2cm]{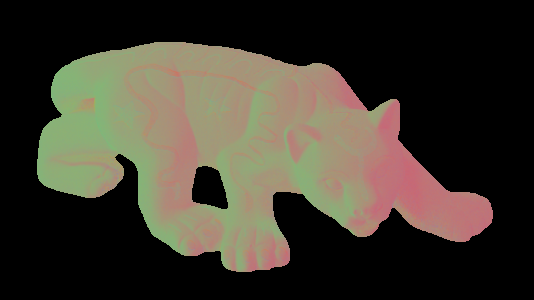}}}
&{{\includegraphics[width=1.8cm,height=1.2cm]{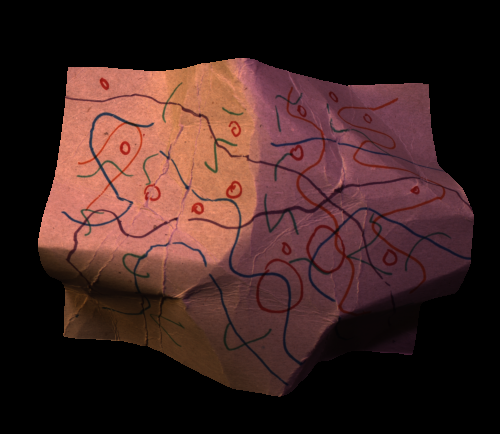}}}
&{{\includegraphics[width=1.8cm,height=1.2cm]{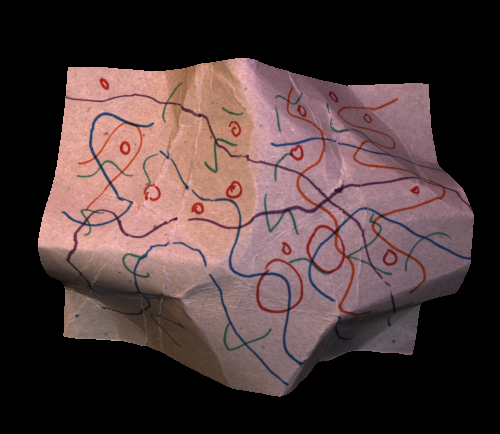}}}
&{{\includegraphics[width=1.8cm,height=1.2cm]{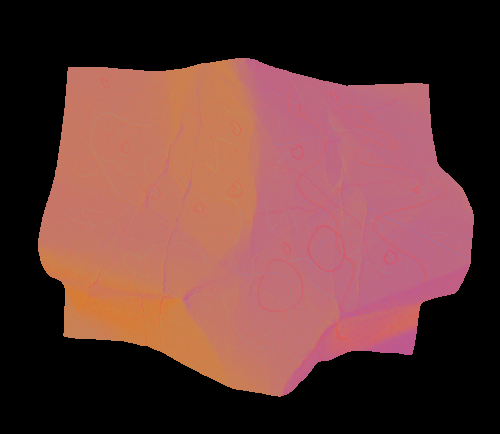}}}
\\
\vspace{-0.75mm}
\raisebox{2\height}{\rotatebox{0}{{ 4 }}}
&{{\includegraphics[width=1.8cm,height=1.2cm]{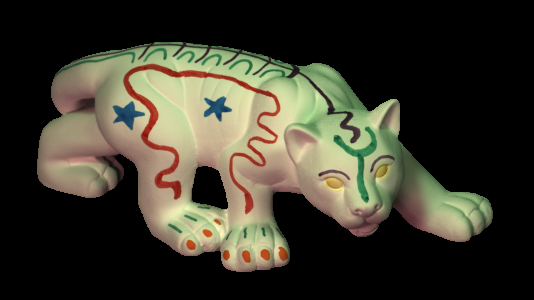}}}
&{{\includegraphics[width=1.8cm,height=1.2cm]{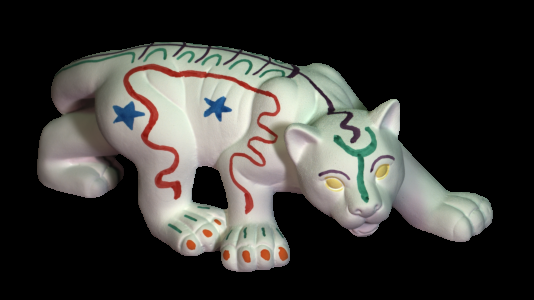}}}
&{{\includegraphics[width=1.8cm,height=1.2cm]{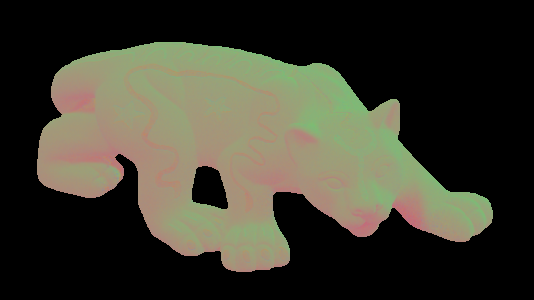}}}
&{{\includegraphics[width=1.8cm,height=1.2cm]{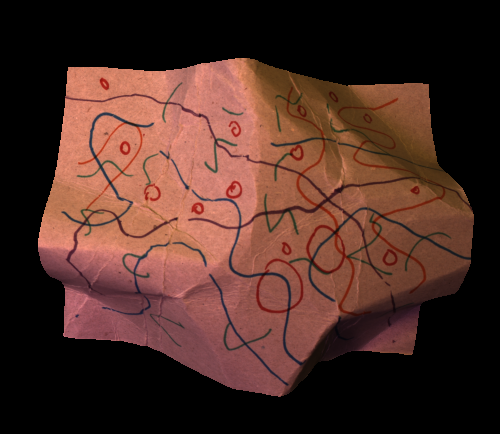}}}
&{{\includegraphics[width=1.8cm,height=1.2cm]{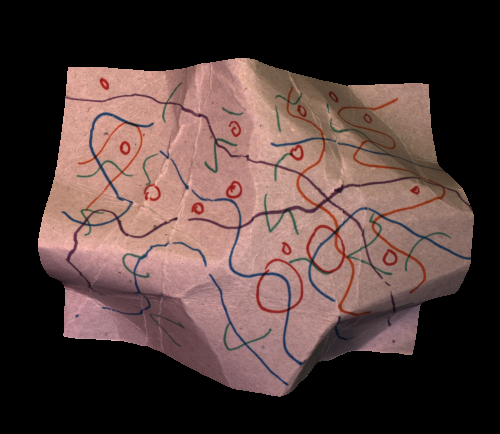}}}
&{{\includegraphics[width=1.8cm,height=1.2cm]{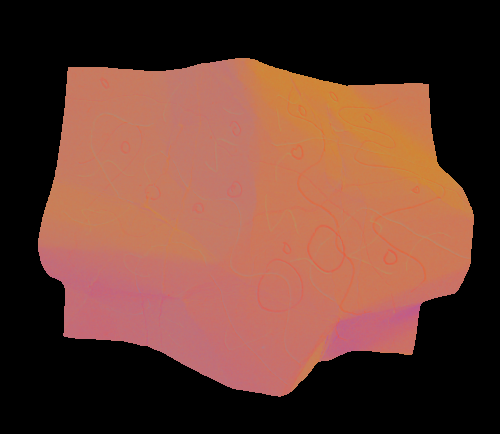}}}
\\
\vspace{-0.75mm}
\raisebox{2\height}{\rotatebox{0}{{ 8 }}}
&{{\includegraphics[width=1.8cm,height=1.2cm]{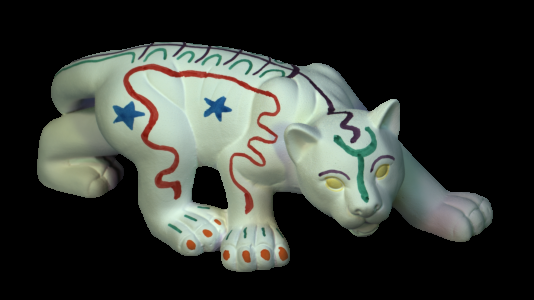}}}
&{{\includegraphics[width=1.8cm,height=1.2cm]{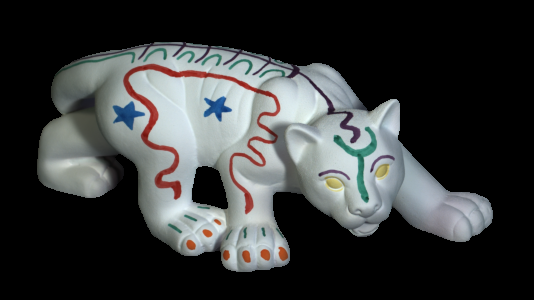}}}
&{{\includegraphics[width=1.8cm,height=1.2cm]{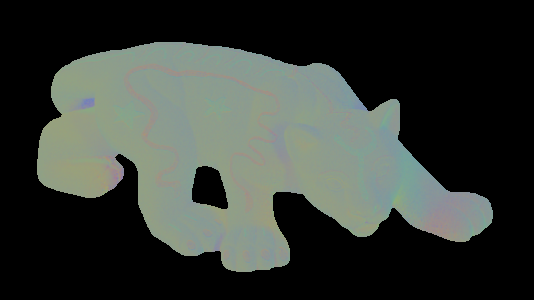}}}
&{{\includegraphics[width=1.8cm,height=1.2cm]{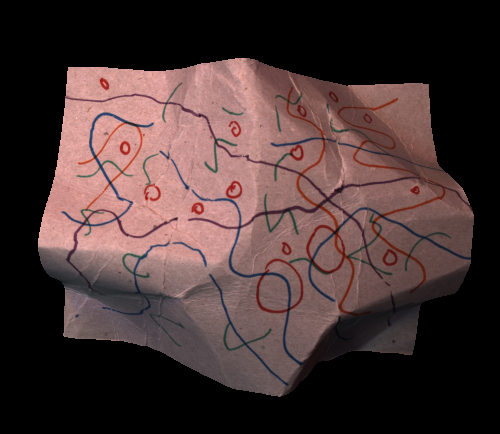}}}
&{{\includegraphics[width=1.8cm,height=1.2cm]{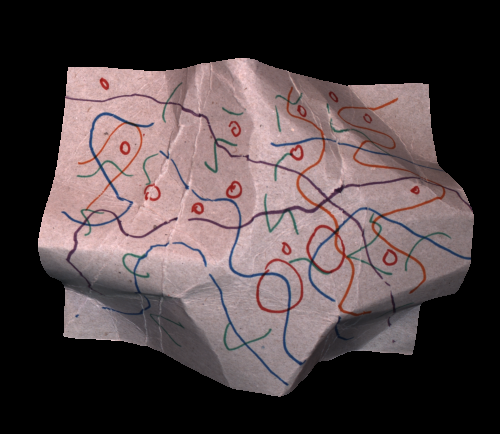}}}
&{{\includegraphics[width=1.8cm,height=1.2cm]{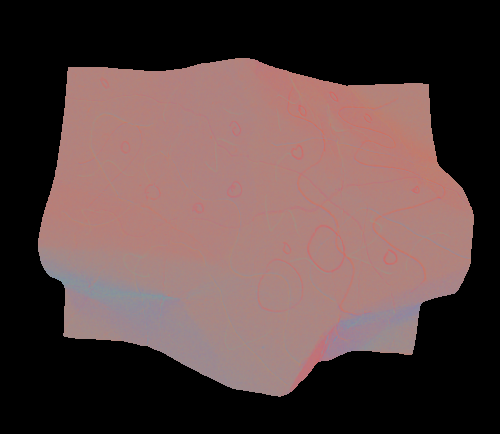}}}
\\

\vspace{-3mm}
\raisebox{2\height}{\rotatebox{0}{{ N }}}
& \raisebox{2\height}{$I$}
& \raisebox{2\height}{$I_f$}
& \raisebox{2\height}{\textit{illu}}
& \raisebox{2\height}{$I$}
& \raisebox{2\height}{$I_f$}
& \raisebox{2\height}{\textit{illu}}

\end{tabular}

\end{center}
\vspace{-2.5mm}
\caption{Example images from the generated flash photography dataset. Each sample is a  triplet: \{no-flash image $I$, flash-on image $I_f$, illumination ground truth map\}. Images are generated from the MIT intrinsic dataset by changing the colors and mixing the original directional illumination images.}
\label{figure:redo_mit}
\vspace{-3mm}
\end{figure*}

\vspace{\medskipamount}\noindent\textbf{Flash gray pixels} can be found from the
grayness map of a residual flash/no-flash image. To select flash gray pixels
robustly, we follow~\cite{yang2015efficient} to compute flash-only illumination component for each pixel:
K-means is used to cluster the top $N\%$ gray pixels into preset $M$ clusters; then the illumination at location $p$ is computed using: 
\vspace{-3mm}
\begin{equation}
\begin{split}
\label{eq:kmean}
L_{fo}^c(p)=\sum_{m=1}^{M} \omega_m L^c_m,
\end{split}
\end{equation}
where $L_m$ refers to the average illumination for the cluster $m$ and $\omega_m$ controls the connection between the pixel $I(x,y)$ to the cluster $m$, unfolded as:
\vspace{-3mm}
\begin{equation}
\begin{split}
\label{eq:d_kmean}
\omega_m=\frac{e^{-\frac{D_m}{2\sigma^2}}}{\sum_{n=1}^{M} e^{-\frac{D_n}{2\sigma^2}}} \enspace , 
\end{split}
\end{equation}
where $D_m$ is the Euclidean distance from the pixel to the centroid of the cluster $m$. Eq.~\ref{eq:d_kmean} encourages nearby pixels to share a similar illumination. 

Combining the flash-only illumination $L_{fo}^c(p)$ with Eq.~\ref{eq:flashimage_formation} allows to color-corrected the image $I_{fo}$
by 
\begin{align}
I_\textit{gray}^{c}(p) = I_\textit{fo}^{c}(p) / L_\textit{fo}^c(p),
\end{align}
and the mixed illumination is:
\begin{align}
L_\textit{}^{c}(p) = I_\textit{}^{c}(p) / I_\textit{gray}^c(p) \enspace .
\end{align}

Flash gray pixel can be filled by more advanced gray pixel methods for further improvement. 

\section{Dataset}
\label{sec:dataset}

We adapted the MIT Intrinsic benchmark~\cite{grosse2009ground} for our task~\footnote{There are recent datasets that provide flash/no-flash image pairs~\cite{Aksoy-2018-CVPR}, but these are unsuitable for our purposes due to unknown illumination number.}. This dataset was originally collected for the intrinsic image decomposition task, containing $20$ single objects illuminated by uncalibrated whitish light sources from $10$ different directions. This property allows us to render each image in the combinations of $1-9$ directed lights with arbitrary chroma to compose a new no-flash image $I$. The $6$-{th} direction is
always roughly frontal and was thus used as the flash source and together with $I$ forms the flash image $I_f$. We generated $I$ and $I_f$ with varying number $N$ of light sources, from $2$ to $8$. Note that even for the easiest case ($N=2$), a pixel may be simultaneously affected by two light sources, violating the global illumination assumption and breaking the original gray pixel as demonstrated in Fig. \ref{figure:redo_mit}.

Considering the fact that gray pixel methods are designed for realistic consumer images "in the wild"~\cite{yang2015efficient} that contain at least a few gray pixels, we
left out the $5$ chromatic objects\footnote{``apple, pear, frog2, potato, turtle''}. In total, the new dataset contains $105$  flash/no-flash image pairs with spatial illumination map ground truth (($15$ objects and $7$ choices of $N$).

\section{Experiments}
\label{sec:results}

\begin{table*}[t]
\begin{center}
\caption{Results for the gray pixel (GP) variants and their flash photography versions with the mixed illumination dataset. The values are  angular errors averaged over image between the estimated and ground truth illumination maps (lower is better). Gray denotes the median and white the mean error. $N$ is the number of light sources. The ``all'' column shows the mean statistics over all choices of $N$. } 
\label{tab:mit}
  \begin{tabular}{l ab ab ab ab ab ab ab ab}
  \hline
 Method/N &  \multicolumn{2}{c}{2} &  \multicolumn{2}{c}{3} &  \multicolumn{2}{c}{4} &  \multicolumn{2}{c}{5} & 
 \multicolumn{2}{c}{6} &  \multicolumn{2}{c}{7} & \multicolumn{2}{c}{8} &  \multicolumn{2}{c}{all} \\
 \toprule
GP \cite{yang2015efficient} & 5.86 & 6.15 & 5.33 & 6.65  & 4.01 & 5.74 & 3.73 & 5.34 & 3.04 & 4.93 & 3.59 & 4.80  & 3.37 & 4.80 & 4.07 & 5.49\\
GP+\textit{f}& 2.37 & 3.49 & 2.32 & 3.85  & 2.35 & 3.79 & 2.39 & 3.94 & 2.39 & 3.80 & 2.39 & 3.85 & 2.39 & 3.88 & 2.37 & 3.80\\
\midrule
MSGP \cite{yanlin2019vissap} & 4.91 & 5.66 & 5.25 & 6.03  & 4.10 & 5.23 & 3.52 & 4.81 & 3.17 & 4.42 & 3.54 & 4.35  & 2.87 & 4.52 & 4.10 & 5.00\\
MSGP +\textit{f}             & 2.34 & 3.17 & 2.29 & 3.32  & 2.34 & 3.31 & 2.34 & 3.40 & 2.33 & 3.38 & 2.50 & 3.72  & 2.30 & 3.46 & 2.34 & 3.40\\
\midrule
DGP \cite{yanlin2019arxiv} & 6.03 & 6.26 & 5.26 & 6.72  & 4.52 & 5.89 & 4.08 & 5.47 & 3.42 & 5.09 & 3.97 & 55.07  & 3.50 & 4.99 & 4.13 & 5.64\\
DGP +\textit{f} & 2.37 & 3.73 & 2.34 & 4.00  & 2.37 & 3.96 & 2.39 & 4.01 & 2.40 & 3.93 & 2.40 & 4.02  & 2.39 & 3.99 & 2.37 & 3.93\\
\bottomrule
  \end{tabular}
\end{center}
\end{table*}

\begin{figure}
\vspace{-10mm}
\begin{center}
\hspace{-28pt}
\begin{tabular}{c cc c c}
\vspace{-0.8mm}
\raisebox{2\height}{\rotatebox{0}{{ \textit{ambi} }}}
&{{\includegraphics[width=1.8cm,height=1.2cm]{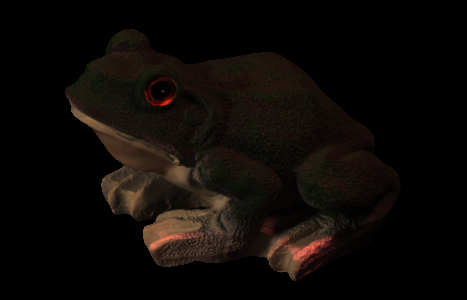}}}
&{{\includegraphics[width=1.8cm,height=1.2cm]{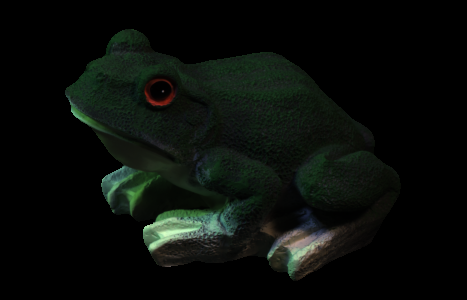}}}
&{{\includegraphics[width=1.8cm,height=1.2cm]{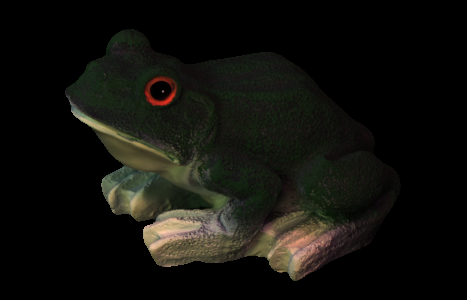}}}
&{{\includegraphics[width=1.8cm,height=1.2cm]{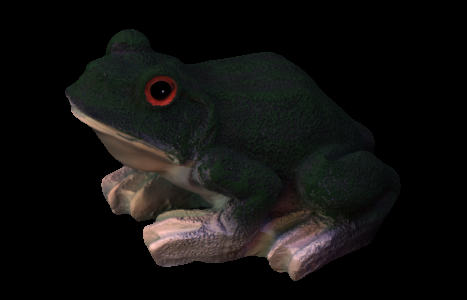}}}

\\
\vspace{-0.8mm}
\raisebox{2\height}{\rotatebox{0}{{ \textit{+flash} }}}
&{{\includegraphics[width=1.8cm,height=1.2cm]{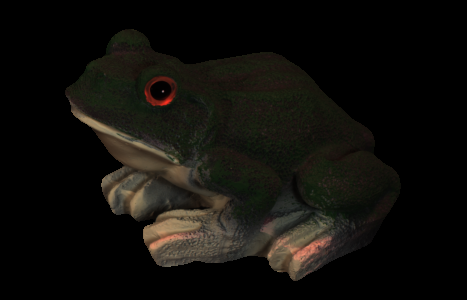}}}
&{{\includegraphics[width=1.8cm,height=1.2cm]{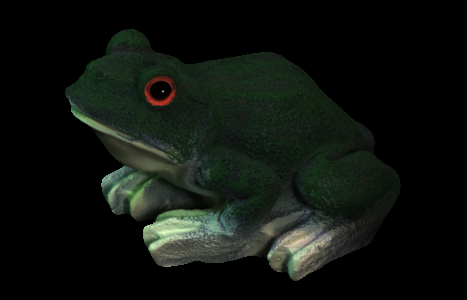}}}
&{{\includegraphics[width=1.8cm,height=1.2cm]{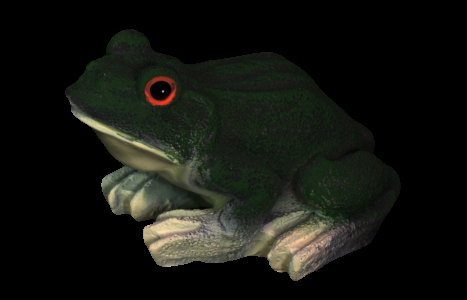}}}
&{{\includegraphics[width=1.8cm,height=1.2cm]{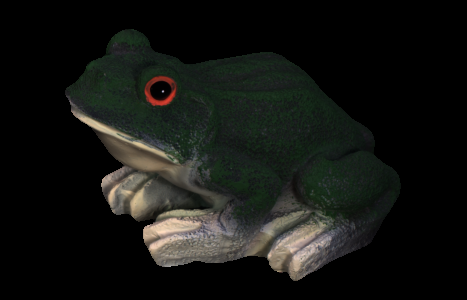}}}

\\
\vspace{-0.8mm}
\raisebox{2\height}{\rotatebox{0}{{ GP }}}
&{{\includegraphics[width=1.8cm,height=1.2cm]{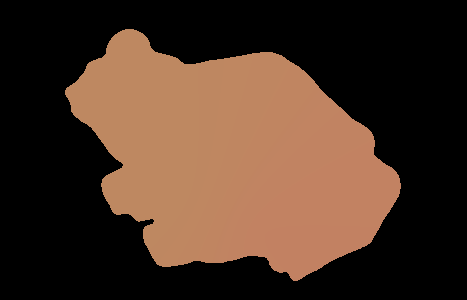}}}
&{{\includegraphics[width=1.8cm,height=1.2cm]{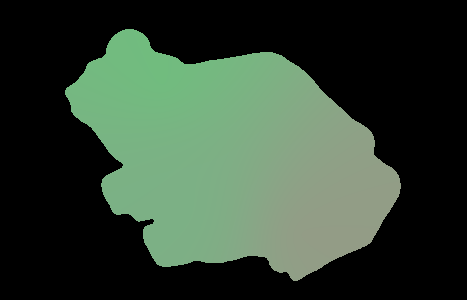}}}
&{{\includegraphics[width=1.8cm,height=1.2cm]{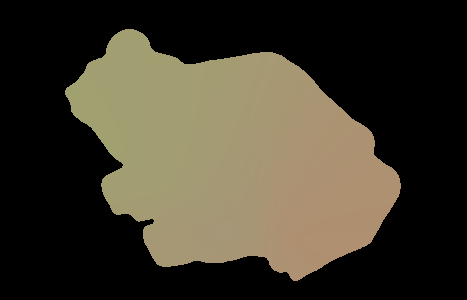}}}
&{{\includegraphics[width=1.8cm,height=1.2cm]{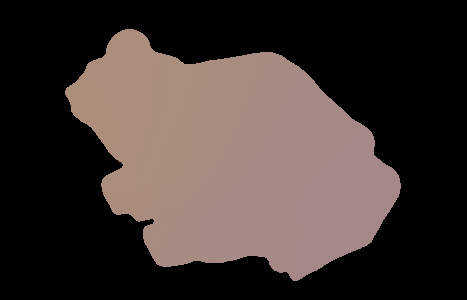}}}

\\
\vspace{-0.8mm}
\raisebox{2\height}{\rotatebox{0}{{ GP+\textit{f} }}}
&{{\includegraphics[width=1.8cm,height=1.2cm]{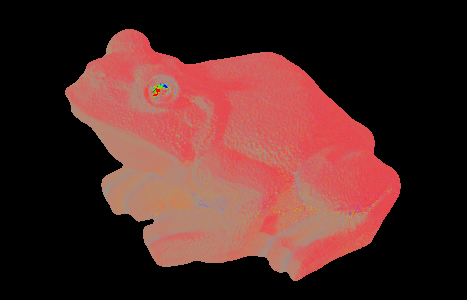}}}
&{{\includegraphics[width=1.8cm,height=1.2cm]{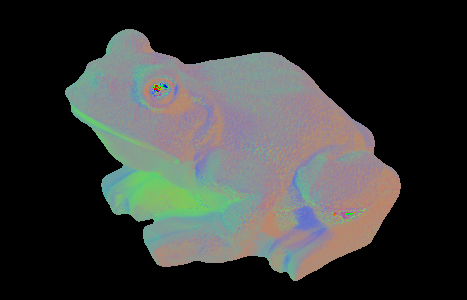}}}
&{{\includegraphics[width=1.8cm,height=1.2cm]{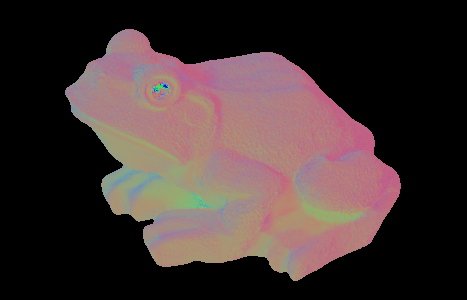}}}
&{{\includegraphics[width=1.8cm,height=1.2cm]{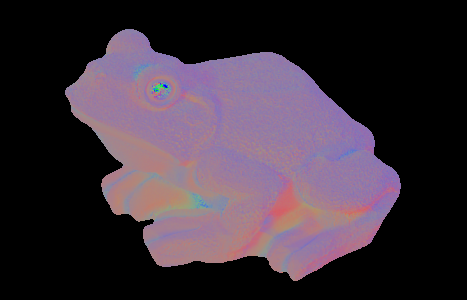}}}
\\
\vspace{-0.8mm}
\raisebox{2\height}{\rotatebox{0}{{ MSGP }}}
&{{\includegraphics[width=1.8cm,height=1.2cm]{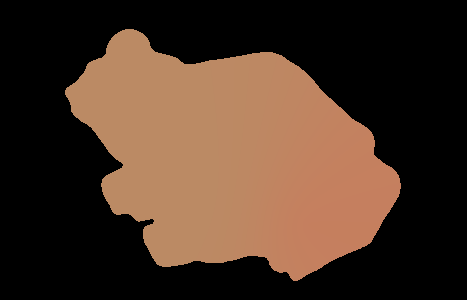}}}
&{{\includegraphics[width=1.8cm,height=1.2cm]{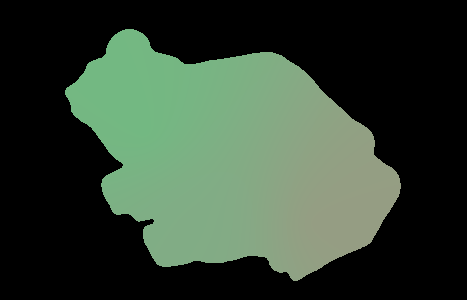}}}
&{{\includegraphics[width=1.8cm,height=1.2cm]{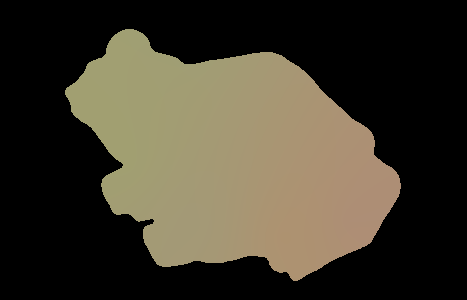}}}
&{{\includegraphics[width=1.8cm,height=1.2cm]{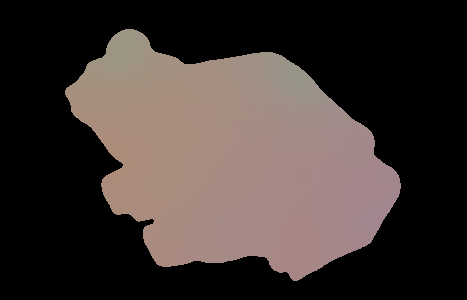}}}
\\
\vspace{-0.8mm}
\raisebox{2\height}{\rotatebox{0}{{ MSGP+\textit{f}  }}}
&{{\includegraphics[width=1.8cm,height=1.2cm]{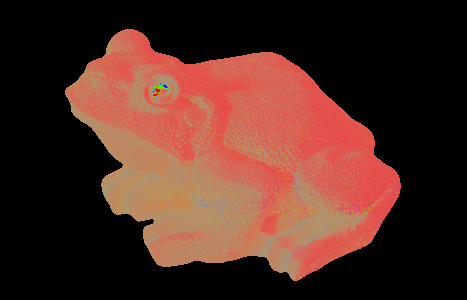}}}
&{{\includegraphics[width=1.8cm,height=1.2cm]{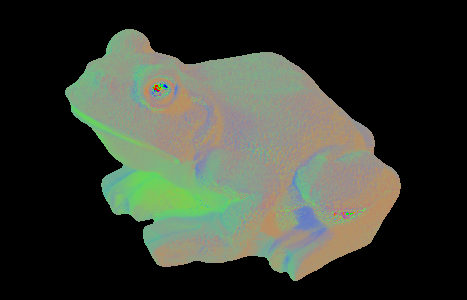}}}
&{{\includegraphics[width=1.8cm,height=1.2cm]{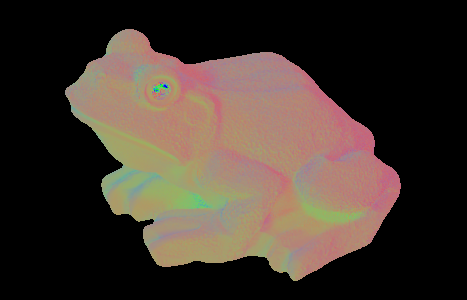}}}
&{{\includegraphics[width=1.8cm,height=1.2cm]{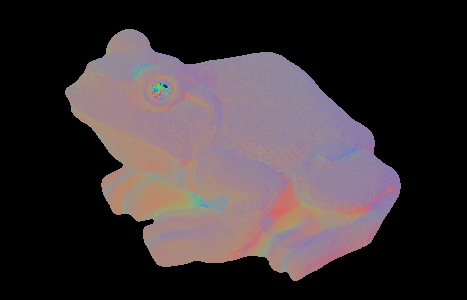}}}
\\
\vspace{-0.8mm}
\raisebox{2\height}{\rotatebox{0}{{ DGP  }}}
&{{\includegraphics[width=1.8cm,height=1.2cm]{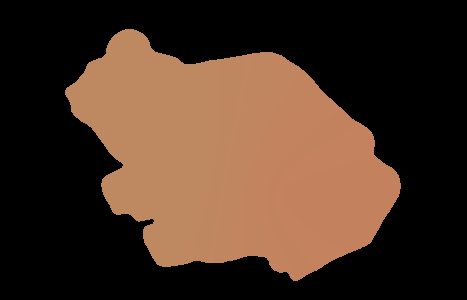}}}
&{{\includegraphics[width=1.8cm,height=1.2cm]{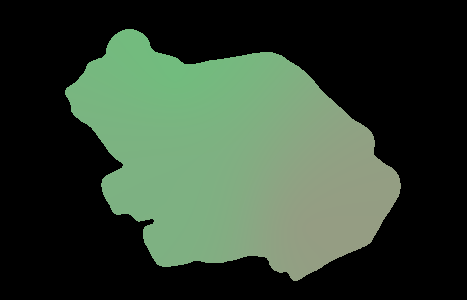}}}
&{{\includegraphics[width=1.8cm,height=1.2cm]{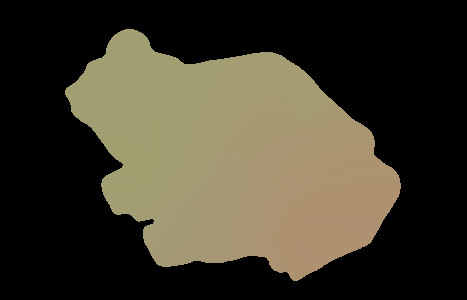}}}
&{{\includegraphics[width=1.8cm,height=1.2cm]{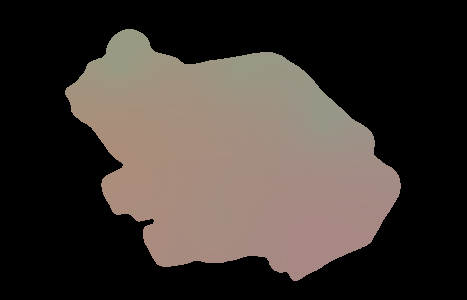}}}
\\
\vspace{-0.8mm}
\raisebox{2\height}{\rotatebox{0}{{ DGP+\textit{f}  }}}
&{{\includegraphics[width=1.8cm,height=1.2cm]{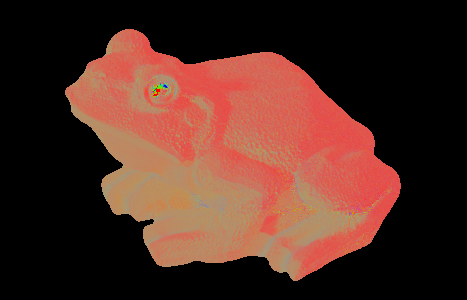}}}
&{{\includegraphics[width=1.8cm,height=1.2cm]{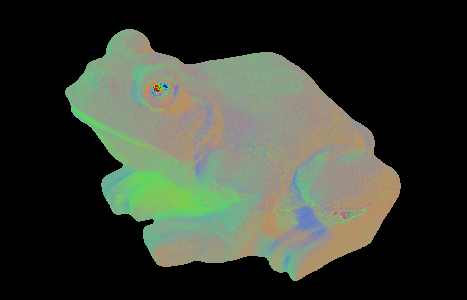}}}
&{{\includegraphics[width=1.8cm,height=1.2cm]{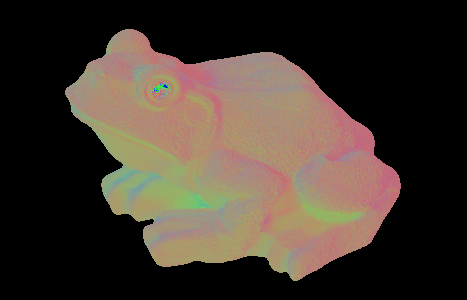}}}
&{{\includegraphics[width=1.8cm,height=1.2cm]{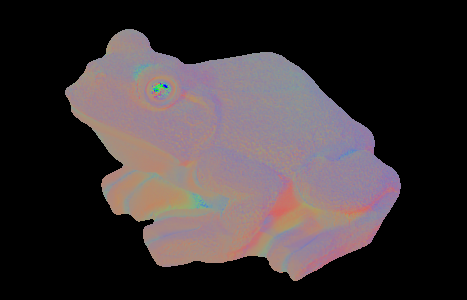}}}
\\

\vspace{-0.8mm}
\raisebox{2\height}{\rotatebox{0}{{ GT }}}
&{{\includegraphics[width=1.8cm,height=1.2cm]{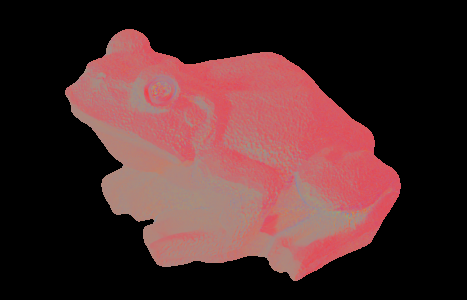}}}
&{{\includegraphics[width=1.8cm,height=1.2cm]{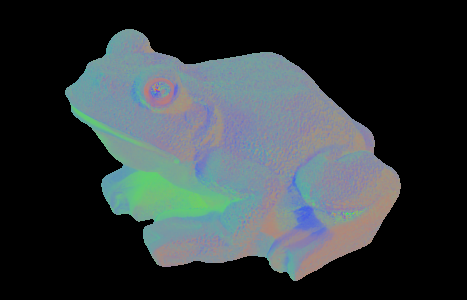}}}
&{{\includegraphics[width=1.8cm,height=1.2cm]{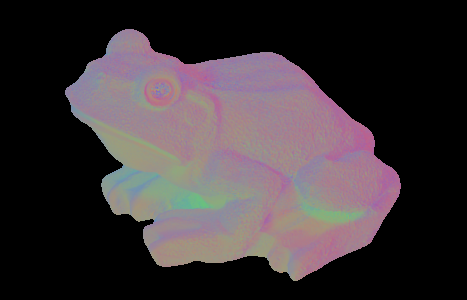}}}
&{{\includegraphics[width=1.8cm,height=1.2cm]{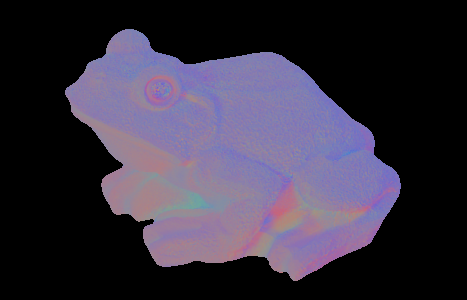}}}

\\
\vspace{-3mm}
\raisebox{2\height}{\rotatebox{0}{{ Method/N }}}
& \raisebox{2\height}{$2$}
& \raisebox{2\height}{$3$}
& \raisebox{2\height}{$5$}
& \raisebox{2\height}{$8$}
\end{tabular}
\end{center}

\vspace{-8mm}
\caption{Qualitative comparison on estimated illumination of flash gray pixels, with $N$ light sources (\textit{2,3,5,8}). }
\label{figure:results_mit}
\vspace{-3mm}
\end{figure}

\setlength{\tabcolsep}{1pt}
\renewcommand{\arraystretch}{1}
\begin{figure}
\begin{center}
\vspace{-0.75mm}
\begin{tabular}{c cc cc}

\hspace{-15pt}
\vspace{-0.75mm}
\raisebox{2\height}{\rotatebox{0}{{  }}}
&{{\includegraphics[width=2cm,height=1.4cm]{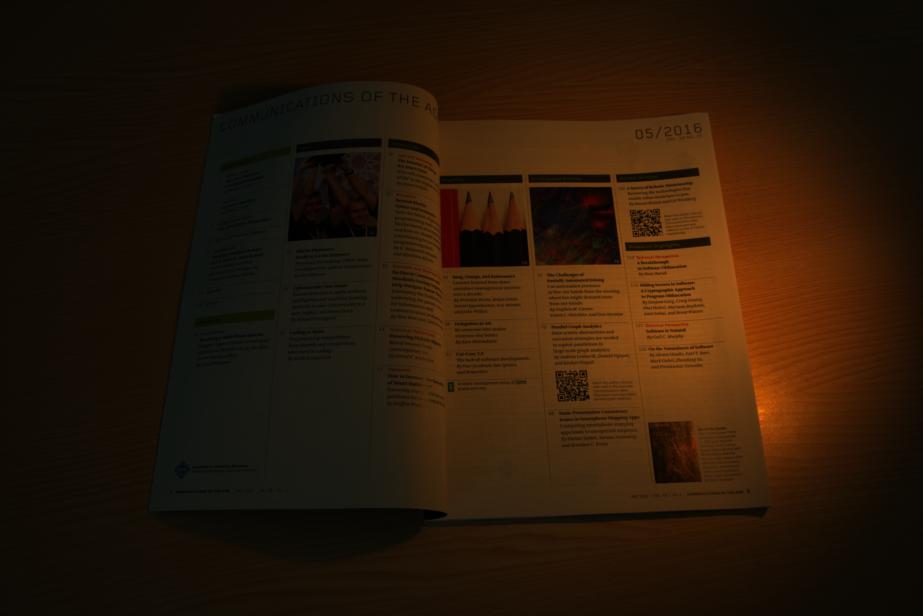}}}
& {{\includegraphics[width=2cm,height=1.4cm]{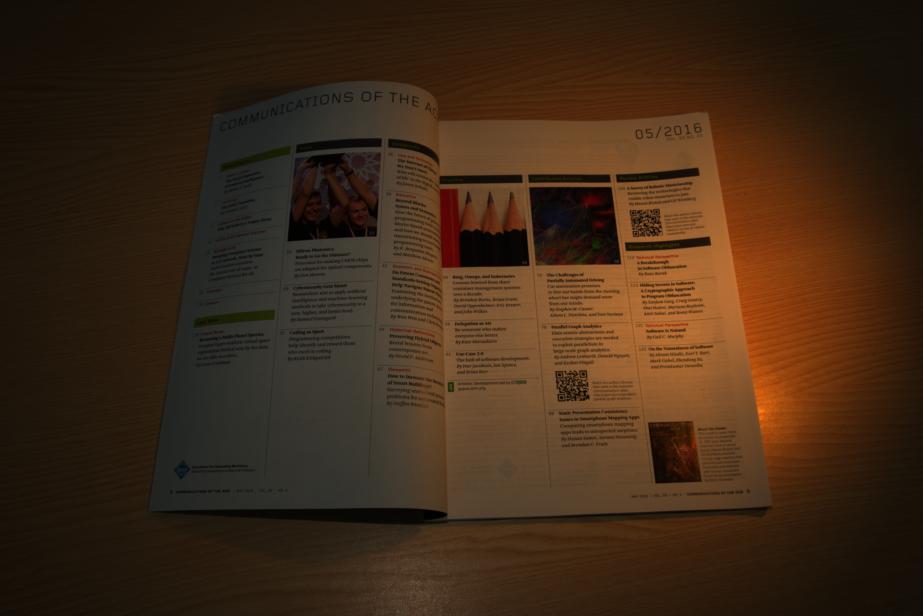}}}
&{{\includegraphics[width=2cm,height=1.4cm]{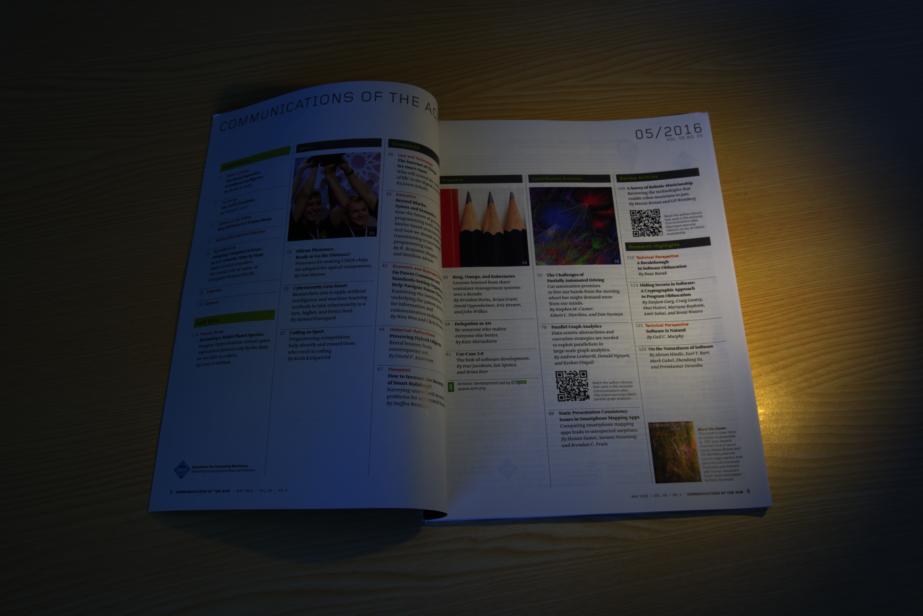}}}
& {{\includegraphics[width=2cm,height=1.4cm]{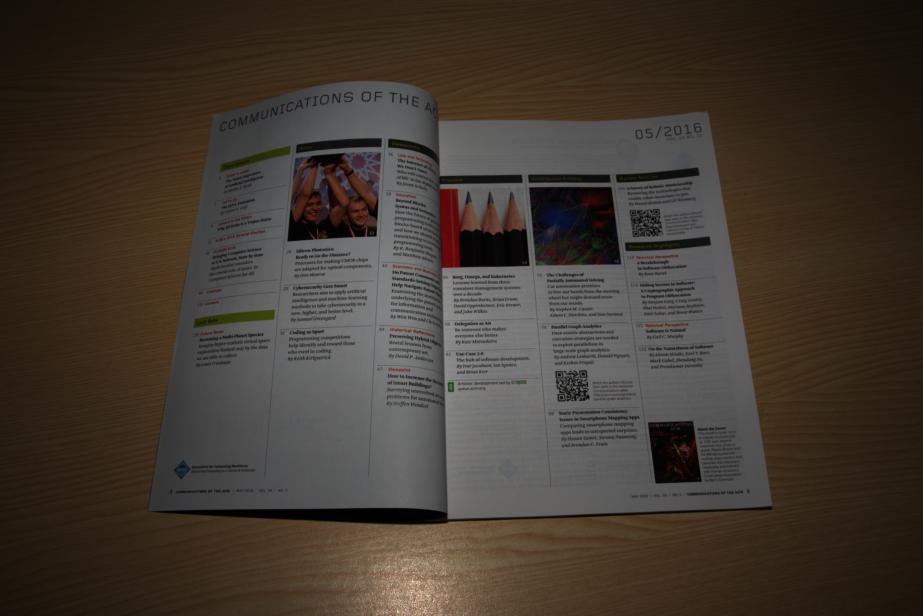}}}

\\

\hspace{-15pt}
\vspace{-0.75mm}
\raisebox{2\height}{\rotatebox{0}{{  }}}
&{{\includegraphics[width=2cm,height=1.4cm]{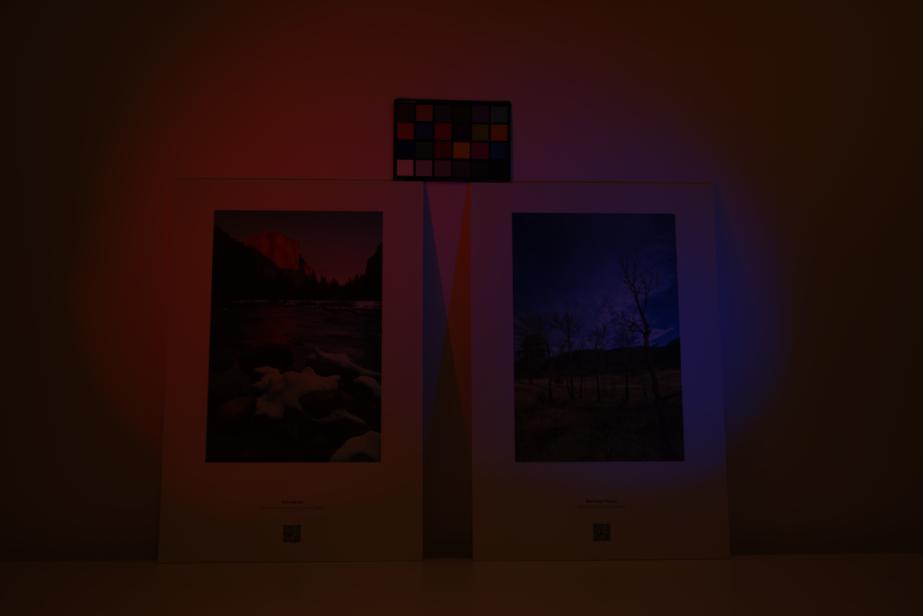}}}
& {{\includegraphics[width=2cm,height=1.4cm]{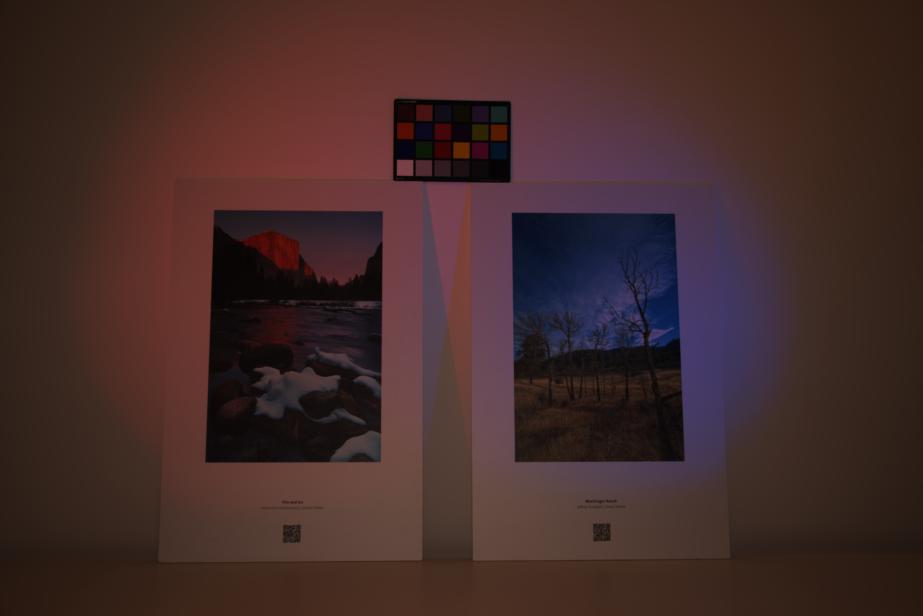}}}
&{{\includegraphics[width=2cm,height=1.4cm]{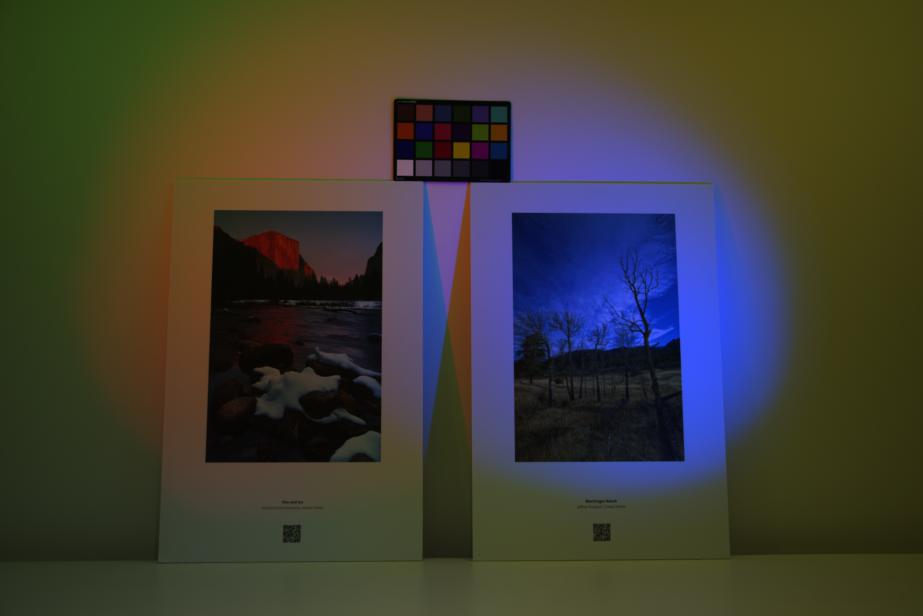}}}
& {{\includegraphics[width=2cm,height=1.4cm]{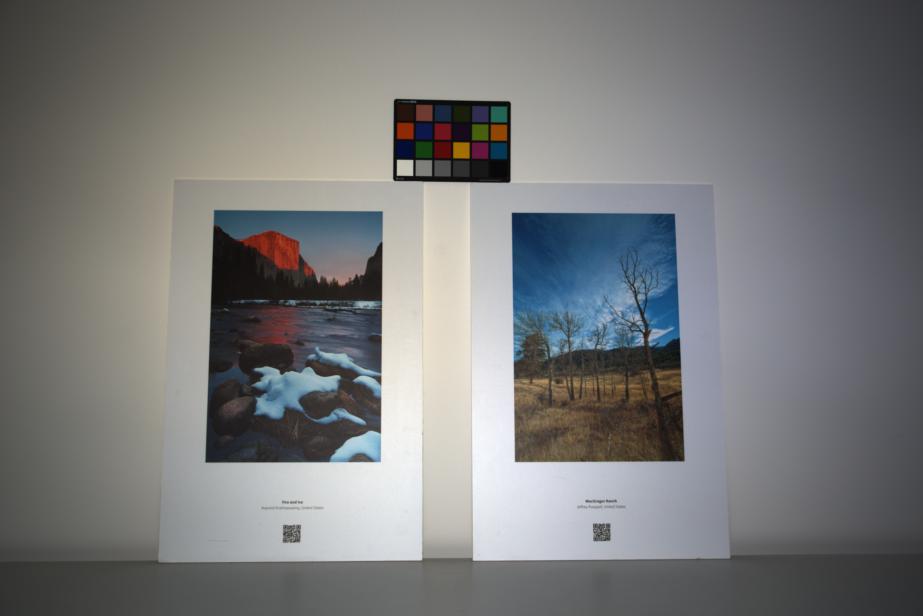}}}

\\

\vspace{-3mm}
\raisebox{2\height}{\rotatebox{0}{{  }}}
& \raisebox{2\height}{\textit{ambi}}
& \raisebox{2\height}{\textit{+flash}}
& \raisebox{2\height}{\textit{GP}}
& \raisebox{2\height}{\textit{GP+f}}

\end{tabular}

\end{center}
\vspace{-8mm}
\caption{
Qualitative comparison on color-corrected image of flash gray pixels in real-world images. Macbeth ColorChecker is excluded for illumination estimation.
}
\label{figure:realimage}
\vspace{-3mm}
\end{figure}

\noindent\textbf{Our setup} is the following: we run the three variants of Gray Pixel methods with flash/no-flash image pairs. On each image the top $10\%$ gray pixels are selected to cover enough area. The cluster number is set to $M=N$, which is the number of illuminants. 

\vspace{\medskipamount}\noindent\textbf{Evaluation metric} is the standard
average angular error~\cite{yang2015efficient}. 

\vspace{\medskipamount}\noindent\textbf{Results} are summarized in
Table~\ref{tab:mit} which shows the performance of flash gray pixel on the dataset in Section~\ref{sec:dataset}. ``GP\textit{+f}'' refers to flash gray pixel on the basis of the original gray pixel method ``GP''. Results show that flash photography extensions
of all gray pixel variants~\cite{yang2015efficient,yanlin2019vissap,yanlin2019arxiv}
systematically improve the results. For all three methods, the improvement over the whole dataset is over $40\%$ in median and $30\%$ in mean. The flash photography
variants achieve the sufficient color constancy accuracy ($\le 3.0^\circ$) in
almost all cases. Fig.~\ref{figure:results_mit} illustrates predicted illumination between gray pixel methods and their flash variants. It is clear that the
original versions cannot find the fine-grained details of mixed illumination.
For example, the frog back and stomach are cast with different colors that
confuses the original GP methods.

Among all gray pixel methods, the original GP \cite{yang2015efficient} suffers from the mixed illumination the most. MSGP \cite{yanlin2019vissap} and DGP \cite{yanlin2019arxiv} perform slightly better, but not to a satisfying degree without
the flash. With flash gray pixel, all three methods perform similarly thus
verifying the efficiency of flash photography. DGP \cite{yanlin2019arxiv} does
not perform better than the original GP which is due to the fact that our
dataset does not contain specular reflectance components.

As the number of diverse illumiants increases, all GP methods performs better.
This can be explained by the fact
that a large number of lights with various colors additively mix toward a
whitish color. Flash gray pixel variants are not affected by this, obtaining consistent results. 

Images from real-world scenes (Fig. \ref{figure:realimage}, images retrieved from \cite{hui2016white}), show that the proposed methods also produce high quality results (\textit{e.g.} the white wall is white).

\section{CONCLUSION}
\label{sec:conclusion}

In this paper we reconsider gray pixel detection through the medium of flash photography. We find that computing a residual map from the flash/no-flash pair, gray pixel methods can effectively measure the grayness of each pixel and provide spatial color constancy.
 We find that computing a residual map from the flash/no-flash pair, gray pixel methods can effectively measure the grayness of each pixel, allowing a large margin improvement in spatially-varying illumination estimation.  The method is pragmatic -- it is lightweight,  does not
need any illuminant prior, training, flash light calibration or
user input.





\clearpage
\bibliographystyle{IEEEbib}
\bibliography{icip2019}

\end{document}